\documentclass[conference]{IEEEtran}
\IEEEoverridecommandlockouts
\usepackage{cite}
\usepackage[colorlinks,citecolor=blue,urlcolor=blue,bookmarks=false,hypertexnames=true]
{hyperref}
\usepackage{amsmath,amssymb,amsfonts}
\usepackage{algorithmic}
\usepackage{graphicx}
\usepackage{textcomp}
\usepackage{xcolor}
\usepackage{caption}
\usepackage{float}
\usepackage{tikz}

\newcommand\copyrighttext{%
  \footnotesize \textcopyright 2026 IEEE. Personal use of this material is permitted.
  Permission from IEEE must be obtained for all other uses, in any current or future
  media, including reprinting/republishing this material for advertising or promotional
  purposes, creating new collective works, for resale or redistribution to servers or
  lists, or reuse of any copyrighted component of this work in other works.
}

\newcommand\copyrightnotice{%
\begin{tikzpicture}[remember picture,overlay]
\node[anchor=south,yshift=10pt] at (current page.south) {\fbox{\parbox{\dimexpr\textwidth-\fboxsep-\fboxrule\relax}{\copyrighttext}}};
\end{tikzpicture}%
}

\def\BibTeX{{\rm B\kern-.05em{\sc i\kern-.025em b}\kern-.08em
    T\kern-.1667em\lower.7ex\hbox{E}\kern-.125emX}}
\begin{document}

\title{MonoSIM: An open source SIL framework for Ackermann Vehicular Systems with Monocular Vision}

\makeatletter
\newcommand{\linebreakand}{%
  \end{@IEEEauthorhalign}
  \hfill\mbox{}\par
  \mbox{}\hfill\begin{@IEEEauthorhalign}
}
\makeatother

\author{\IEEEauthorblockN{Shantanu Rahman*}
\IEEEauthorblockA{\textit{Department of Electrical and} \\ \textit{Electronic Engineering} \\
\textit{Islamic University of Technology}\\
Gazipur, Bangladesh \\
shantanurahman@iut-dhaka.edu}
*Corresponding author
~\\
\and
\IEEEauthorblockN{Nayeb Hasin}
\IEEEauthorblockA{\textit{Department of Electrical and} \\ \textit{Electronic Engineering} \\
\textit{Islamic University of Technology}\\
Gazipur, Bangladesh \\
nayebhasin@iut-dhaka.edu}
~\\
\and
\IEEEauthorblockN{Mainul Islam}
\IEEEauthorblockA{\textit{Department of Electrical and} \\ \textit{Electronic Engineering} \\
\textit{Islamic University of Technology}\\
Gazipur, Bangladesh \\
mainulislam@iut-dhaka.edu}
~\\
\linebreakand
\IEEEauthorblockN{Md. Zubair Alom Rony}
\IEEEauthorblockA{\textit{Department of Electrical and} \\ \textit{Electronic Engineering} \\
\textit{Islamic University of Technology}\\
Gazipur, Bangladesh \\
zubairalom@iut-dhaka.edu}

\and
\IEEEauthorblockN{Golam Sarowar}
\IEEEauthorblockA{\textit{Department of Electrical and} \\ \textit{Electronic Engineering} \\
\textit{Islamic University of Technology}\\
Gazipur, Bangladesh \\
asim@iut-dhaka.edu}
}

\maketitle
\copyrightnotice

\begin{abstract}
This paper presents an open-source Software-in-the-Loop (SIL) simulation platform designed for autonomous Ackerman vehicle research and education. The proposed framework focuses on simplicity, while making it easy to work with small-scale experimental setups, such as the XTENTH-CAR platform. The system was designed using open source tools, creating an environment with a monocular camera vision system to capture stimuli from it with minimal computational overhead through a sliding window based lane detection method. The platform supports a flexible algorithm testing and validation environment, allowing researchers to implement and compare various control strategies within an easy-to-use virtual environment. To validate the working of the platform, Model Predictive Control (MPC) and Proportional–Integral–Derivative (PID) algorithms were implemented within the SIL framework. The results confirm that the platform provides a reliable environment for algorithm verification, making it an ideal tool for future multi-agent system research, educational purposes, and low-cost AGV development. Our code is available at \textit{https://github.com/shantanu404/monosim.git}.
\end{abstract}


\section{Introduction}
Autonomous vehicles are quickly becoming a key transportation technology. Companies like Tesla are accelerating development and increasing market demand. In many developed countries, users are beginning to depend on these systems~\cite{r2}. Since human error remains a major cause of road accidents, autonomous driving offers a promising pathway toward improved safety~\cite{r3}. Motivated by these trends, we introduce MonoSIM, a lightweight and accessible simulation testbed designed to support research on autonomous vehicle navigation.

MonoSIM is built on top of the Robot Operating System (ROS) and the MVSIM simulator~\cite{blanco2023multivehicle}, which incorporates the Box2D physics engine. Through this design, MonoSIM enables lane-tracking experiments using monocular vision and widely used control algorithms. The framework provides telemetry extraction, real-time sensor emulation, and a flexible environment for developing low-level vision pipelines—capabilities that are often restricted in traditional control-focused simulators. MonoSIM places emphasis on real-time sensor data processing, dynamic region-of-interest extraction for lane detection, sliding-window techniques, quadratic curve fitting, and improved transparency of the simulation environment. Practical constraints such as limited computational power and low-resolution camera input are also taken into consideration within the system’s design.

Several prior works have proposed testbeds for autonomous systems. Jiang et al.\ integrated an MPC framework with a learning-based error estimator using offline data~\cite{x1}. Bhadani et al.\ developed the “CAT Vehicle Testbed” using ROS and Gazebo to support hardware-in-the-loop experimentation~\cite{x2}. Sarantinoudis et al.\ presented a ROS-based platform supporting V2V and V2I communication~\cite{x3}. Elmoghazy et al.\ proposed an edge-enhanced V2X testbed using an F1Tenth-scale vehicle~\cite{x4}. Bouchemal et al.\ constructed a real-world DSRC-based V2X infrastructure~\cite{x5}. Additional contributions include fault-simulation platforms~\cite{x6}, large-scale heterogeneous testbeds~\cite{x7}, controlled-environment autonomous prototypes~\cite{x8}, reconfigurable multi-sensor platforms~\cite{x9}, radar-focused MATLAB testbeds~\cite{x10}, integrated UAV–UGV environments~\cite{x12}, and digital-twin-supported hybrid testbeds~\cite{x13}.

Compared to these existing platforms, MonoSIM offers lower computational demands and improved scalability for testing and comparing autonomous vehicle control algorithms within a unified monocular-vision-driven simulation environment.

\section{Methodology}
Sensors and locomotion are necessary to make interaction with the environment. In our testbed, we used a monocular camera (\textit{Logitech c310 HD}) to observe the environment around. For actuation, the car model we developed, follows the mathematical model of ackermann steering using an approximation of the bicycle model. Ackermann steering helps to turn a car by adjusting the rotating angle of wheels according to their position where, The bicycle model simplifies the vehicle's dynamics by representing the vehicle with a single front wheel and a single rear wheel.

\subsection{CALIBRATION}
For consistent processing, a frame from the target environment was taken and enlarged to $640 \times 480$ pixels. OpenCV was used to convert the image from BGR to RGB format for viewing.

\begin{figure}[htbp]
    \centering
    \begin{minipage}{0.49\linewidth}
        \centering
        \includegraphics[width=0.9\linewidth]{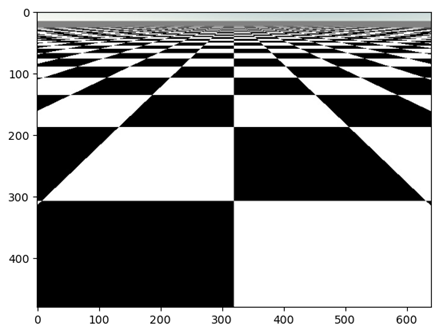}
        \caption{Original input frame}
        \label{Fig:2.1}
    \end{minipage}%
    \hfill%
    \begin{minipage}{0.49\linewidth}
        \centering
        \includegraphics[width=0.9\linewidth]{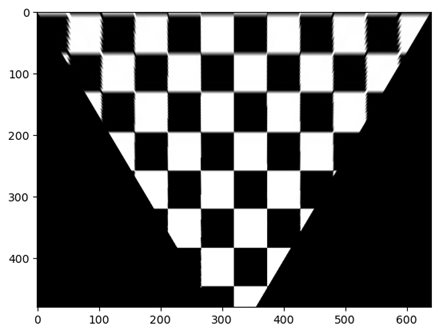}
        \caption{Birds Eye View}
        \label{Fig:2.2}
    \end{minipage}
\end{figure}

A homography-based perspective transformation was used to create a top-down view. After the source and destination points were manually specified, the Direct Linear Transformation algorithm was used to calculate the transformation matrix. After that, the picture was distorted to a corrected viewpoint.

The calibration was performed using a $5 \times 4$ chessboard. Using OpenCV's \texttt{findChessboardCorners} with adaptive thresholding and normalizing for enhanced robustness, chessboard corners were identified after the corrected image was converted to grayscale.

\begin{figure}
    \centering
    \includegraphics[width=0.7\linewidth]{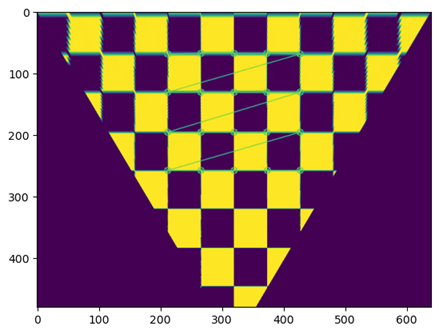}
    \caption{Detected and refined chessboard corners overlaid on the grayscale image.}
    \label{Fig:2.3}
\end{figure}

Corner positions were refined to subpixel accuracy using \texttt{cornerSubPix}, and the results were visualized for verification. The corner coordinates were reshaped to evaluate geometric consistency, and average spacing between adjacent corners was computed in both horizontal and vertical directions.

Finally, the homography matrix was printed and confirmed. This matrix provides a geometric mapping between original and rectified views and is essential for applications such as camera calibration, navigation, and ground plane analysis.

\subsection{LANE DETECTION}
\label{lane-detection}
The feature-based lane detection pipeline uses edges and local visual cues such as gradients and intensity variations, which remain stable across different road geometries but may vary with illumination. The camera output is first resized to $640 \times 480$ and undistorted using the intrinsic camera matrix obtained by camera calibration.

Because fitting polynomials directly on the perspective image would require high-order models, a bird's-eye view transform is applied \cite{r28} to make the lanes parallel and fit them using a second or third-order polynomial (see Fig.~\ref{Fig:2.5} and Fig.~\ref{Fig:2.6}).

\begin{figure}[htbp]
    \centering
    \begin{minipage}{0.24\textwidth}
        \centering
        \includegraphics[width=0.75\linewidth]{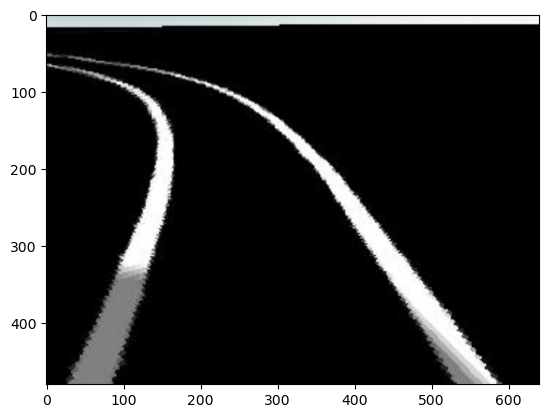}
        \caption{Lane visible in camera}
        \label{Fig:2.5}
    \end{minipage}%
    \hfill%
    \begin{minipage}{0.24\textwidth}
        \centering
        \includegraphics[width=0.75\linewidth]{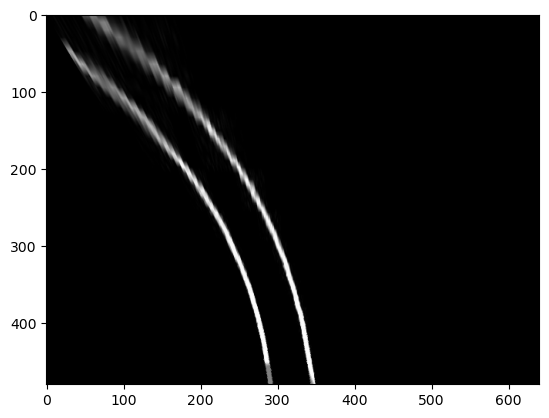}
        \caption{Bird’s Eye view}
        \label{Fig:2.6}
    \end{minipage}

\end{figure}

After warping, the image is converted to HSV, and lane features are isolated primarily from the Value channel. The lower half of this channel is processed using Gaussian blurring, thresholding, and morphological filters. A vertical histogram of pixel intensities is then used to estimate the initial left and right lane positions 

\vspace{-6pt}
\begin{figure}[H]
    \centering
    \includegraphics[scale=0.4]{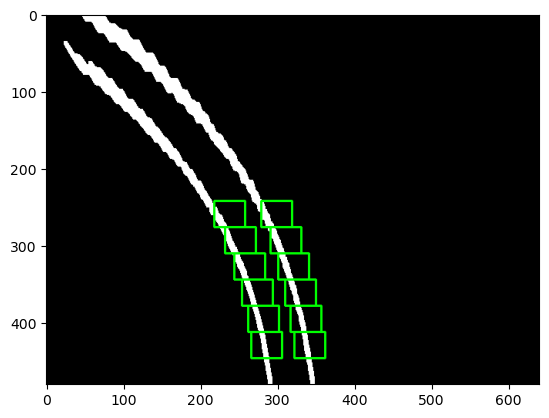}
    \caption{Implementation of Sliding window technique}
    \label{Fig:2.11}
\end{figure}
\vspace{-10pt}

Lane pixels are extracted using a sliding-window method, where windows move upward along the image to collect connected lane points. Separate polynomials are fitted to the left and right lane boundaries, providing a continuous representation of lane geometry. (Fig.~\ref{Fig:2.11}). An overview of the entire pipeline is shown in Fig.~\ref{Fig:2.12}.

\vspace{-6pt}
\begin{figure}[H]
    \includegraphics[width=\linewidth]{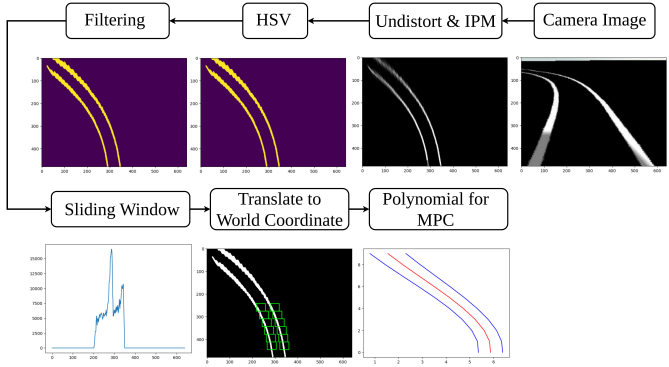}
    \caption{Overall lane detection pipeline}
    \label{Fig:2.12}
\end{figure}
\vspace{-10pt}

\subsection{VEHICLE DYNAMICS}
Before developing our vehicle dynamics model, we are assuming only 3 DOF of motion for our autonomous vehicle i.e. lateral, longitudinal and yaw motion. We are also making some assumption based on 3 DOF motions which are as follows \cite{r24}:
\begin{itemize}
    \item For the front and rear axles, the wheels are considered lumped together.
    \item Assuming fixed velocity for our autonomous vehicle
    \item Maneuvering only the front axel.
    \item The weight distribution is considered uniform throughout the body and extra dynamics such as suspension, aerodynamics, and slip factors are discarded
\end{itemize}
Considering Newton’s Second Law of motion for 3 DOF  \cite{r25}, vehicle dynamics model can be constructed as :
\begin{equation}
m(\ddot{y} + \dot{x} \dot{\phi}) = 2(F_{yf} \sin \delta_f + F_{cf} \cos \delta_f) + 2F_{cr} \label{Eq: 1}
\end{equation}
\begin{equation}
I_z \ddot{\phi} = 2b(F_{yf} \sin \delta_f + F_{cf} \cos \delta_f) - 2a F_{cr} \label{Eq: 2}
\end{equation}
In the above equation Vehicle mass is denoted using m, inertia along z axis is denoted using \(I_z\),  \(\phi\) and \(\delta_f\) are the yaw and steering angle respectively. Distance from the center of gravity to the front and rear axels are denoted using b and a respectively.

Pacejka tire model \cite{r26} is applied in this paper, Where the cornering angle and slip ratio of tire are quite small, the linearized and simplified formulas of longitudinal force and lateral force in the tire model are given by:

\begin{equation}
F_{cf} = C_{cf} \alpha_f \label{Eq: 3}
\end{equation}
\begin{equation}
F_{cr} = C_{cr} \alpha_r \label{Eq: 4}
\end{equation}

where \(C_cf\) and \(C_cr\) are the front and the rear cornering tire stiffness coefficient, respectively. Similarly, \(\alpha_f\) and \(\alpha_r\) are the front and rear tire slip angles, respectively.

Real time performance can be improved and optimized by converting the nonlinear dynamic model into the state space model and linearized, where \(\chi= \dot{x} ,\dot{y} ,\phi, \dot{\phi}^T \) is the state space vector and the control variable is  \(u=\delta_f\) \cite{r27} to increase the versatility of the model and approximate linearized model is formed.

\begin{figure}[H]
    \centering
    \includegraphics[width=0.85\linewidth]{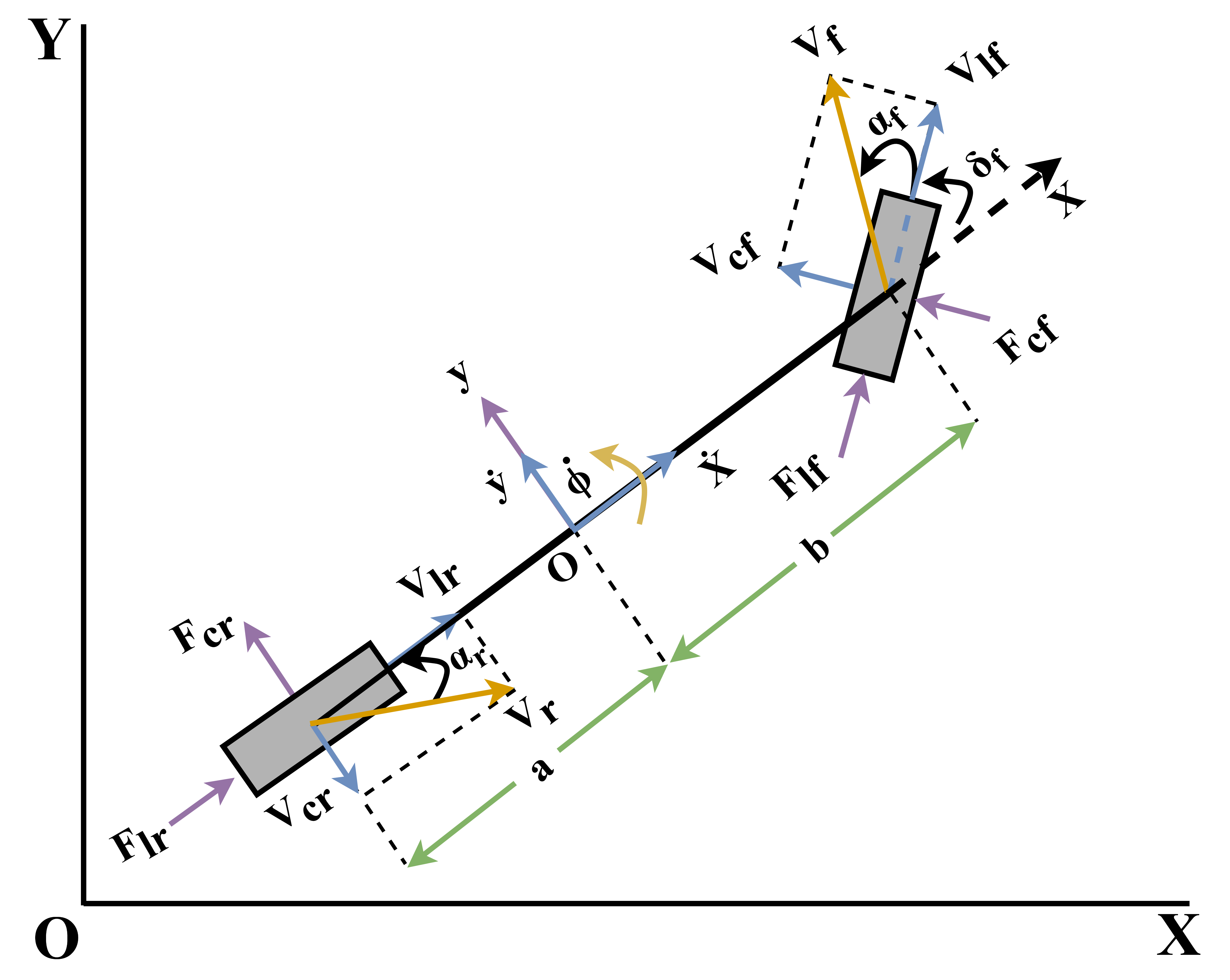}
    \caption{Dynamics of Ackermann model}
    \label{Fig:2.13}
\end{figure}

where
\begin{equation}
A_t = \left. \frac{\partial f(\chi, u)}{\partial \chi} \right|_{\chi = \chi_0; \, u = u_0} \label{Eq: 7}  
\end{equation}
\begin{equation}
B_t = \left. \frac{\partial f(\chi, u)}{\partial u} \right|_{\chi = \chi_0; \, u = u_0} \label{Eq: 8}   
\end{equation}
\begin{equation}
Y_t = \begin{bmatrix} \phi \\ Y \end{bmatrix}C_t = \begin{bmatrix} 1 , 0 , 0 , 0 \end{bmatrix}^T \label{Eq: 9}   
\end{equation}

Due to the continuous nature of the above equation, these equations cannot be applied directly into the model predictive control algorithm, so discretization is performed.

\begin{equation}
\chi(k+1) = A_k \chi(k) + B_k u(k) \label{Eq: 10}
\end{equation}
\begin{equation}
Y(k) = C_k \chi(k) \label{Eq: 11}    
\end{equation}

\subsection{STATE AND OUTPUT PREDICTION}
In the process of path tracking, we need to predict the future behavior of vehicle in the specified prediction horizon, and calculate the control input in the next moment by minimizing the error between predictive variables and the references under various constraints.
Given that the input of the model is the control increment \(\Delta u(k)\) of the control signal \(u(k)\), an incremental state space model is formulated as follows:

\begin{equation}
\tilde{\chi}(k+1) = \tilde{A}_k \tilde{\chi}(k) + \tilde{B}_k \Delta u(k) \label{Eq: 12}
\end{equation}
\begin{equation}
\tilde{Y}(k) = \tilde{C}_k \tilde{\chi}(k) \label{Eq: 13}
\end{equation}

Where the symbols retain their traditional meaning.

To forecast the system's state at future intervals, assuming the current sampling time is \( k \), where \( k > 0 \), and the state variable vector \( \tilde{\chi}(k) \) denotes the present system information. The forthcoming control sequence is represented by \( \Delta u(k + j) = 0 \), where \( j = N_c, N_c + 1, \ldots, N_p - 1 \). \( N_c \) and \( N_p \) denote the control and prediction horizons, respectively.

\begin{equation}
\Delta U_a(k) \triangleq \begin{bmatrix} 
\Delta u(k) \\ 
\Delta u(k+1) \\ 
\vdots \\ 
\Delta u(k+N_c-1) 
\end{bmatrix}_{(N_c \times 1)} \label{Eq: 17}
\end{equation} 

\begin{equation}
\begin{aligned}
\tilde{\chi}(k+1|k) & = \tilde{A}_k \tilde{\chi}(k) + \tilde{B}_k \Delta u(k) \\[1em]
\tilde{\chi}(k+2|k) & = \tilde{A}_{k+1} \tilde{A}_k \tilde{\chi}(k) + \tilde{A}_k \tilde{B}_k \Delta u(k) \\
& \quad + \tilde{B}_{k+1} \Delta u(k+1) \\[1em]
& \vdots \\[1em]
\tilde{\chi}(k+N_c|k) & = \alpha(k,N_c-1,0) \tilde{\chi}(k) \\
& \quad + \alpha(k,N_c-1,1) \tilde{B}_k \Delta u(k) \\
& \quad + \cdots + \tilde{B}_{k+N_c-1} \Delta u(k+N_c-1) \\[1em]
& \vdots \\[1em]
\tilde{\chi}(k+N_p|k) & = \alpha(k,N_p-1,0) \tilde{\chi}(k) \\
& \quad + \alpha(k,N_p-1,1) \tilde{B}_k \Delta u(k) \\
& \quad + \cdots \\
& \quad + \alpha(k,N_p-1,N_c) \tilde{B}_{k+N_c-1} \Delta u(k+N_c-1)
\end{aligned}
\label{Eq: 18}
\end{equation}

The prediction horizon is equal to or, often times, greater than the control horizon. Consequently, the prediction input vector \( \Delta u_a(k) \) can be delineated.

State vector and output of every sample instance in the prediction horizon can be formulated in Eq \ref{Eq: 18}. Using similar technique we can then calculate trajectory for any given set of input controls for the horizons.

\subsection{CONSTRUCTION OF COST FUNCTION}
The fundamental goal of the controller is to determine the sequence of control actions that will minimize the sum of the squared deviations of the expected output from the reference trajectory. This is the basic objective of the controller, the cost function can be defined as:

\begin{equation}
\begin{aligned}
J_k & = \left\lceil \tilde{Y}_{(a,\text{ref})}(k) - \tilde{Y}_a(k) \right\rceil^T Q \left\lceil \tilde{Y}_{(a,\text{ref})}(k) - \tilde{Y}_a(k) \right\rceil \\
& \quad + \Delta U_a^T(k) R \Delta U_a(k)    
\end{aligned}
\end{equation}

The matrices that contain the weights of the inputs and outputs are denoted as \( Q \) and \( R \). The matrix \( \tilde{Y}_{(a,\text{ref})}(k) \) is the reference matrix. For this experiment they have been tuned manually.

\section{Experimental Setup}
The validation of the proposed method is performed on an Ubuntu 24.04 LTS, equipped with a 12 generation Intel Core i7 mobile processor. To make sure we test the algorithms on isolation, we aimed to make the environment as abstract as possible to avoid any bias.

For that reason, we chose MVSIM as our simulation platform which boasts on its 2.5D capable simulation using Box2D engine \cite{blanco2023multivehicle}. ROS2 messaging service was used to communicate between the simulation and the algorithm process. ROS2 was chosen because it provided modularity and scalability to the system. The simulation was run in the above-mentioned setup together with some other ros2 nodes that collected odometry data for comparison with other algorithms. The graph in Fig.~\ref{Fig:2.15} shows all the components that were run in a standard experiment.

The environment was almost black and white with clearly defined lane
lines to indicate the vehicle path. This was in turn detected from the environment using the described method in Section \ref{lane-detection}. This process is analogous to the SOTA ML models segmenting the image into different lane lines. Together with some calibration data from the monocular camera, we were able to deduce the physical points in the actual world. This is again analogous to the case for ML models deducing the depth of pixels in the image. This allowed us to focus only on fine-tuning the underlaying algorithm without being concerned with other external factors.

\begin{figure}[H]
    \centering
    \includegraphics[scale=0.37]{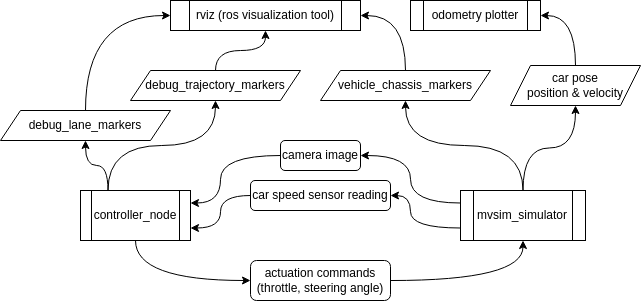}
    \caption{Software Architecture of Simulation}
    \label{Fig:2.15}
\end{figure}

The controller node is responsible for driving the car. It does so by sending the simulator program its desired throttle (accelaration) and steering angle in the form of a ros2 twist message. The controller derives these values from the input camera image and the car's speed using a simulated speedometer sensor located on the vehicle. The controller node then processes this information with the desired algorithm to dictate the throttle/brake and steering angle of the vehicle.

Figure~\ref{Fig:2.17} shows ROS2 \texttt{rviz} showing the monocular camera output along with the detected lane and the predicted path to follow. In Figure \ref{Fig:2.16}, we can see that MVSIM simulated world in isometric view.

Figure~\ref{Fig:SampleRun} shows a sample runtime information of the vehicle while running the simulation. The plots consecutively show its current position in world coordinates, the velocity and the heading angle along with it's angular velocity.

\begin{figure}[!htb]
    \centering
    \includegraphics[width=0.7\linewidth]{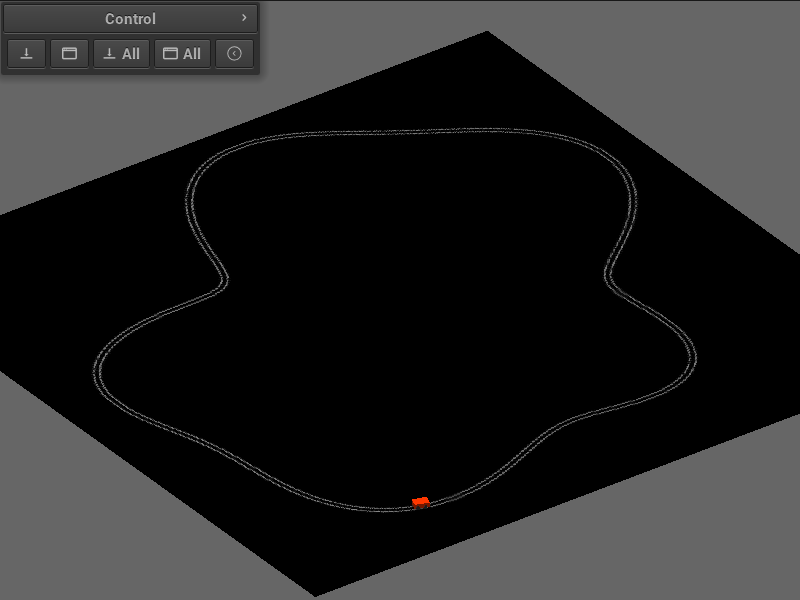}
    \caption{MVSIM Simulation platform}
    \label{Fig:2.16}
\end{figure}

\begin{figure}[!htb]
    \centering
    \includegraphics[width=0.7\linewidth]{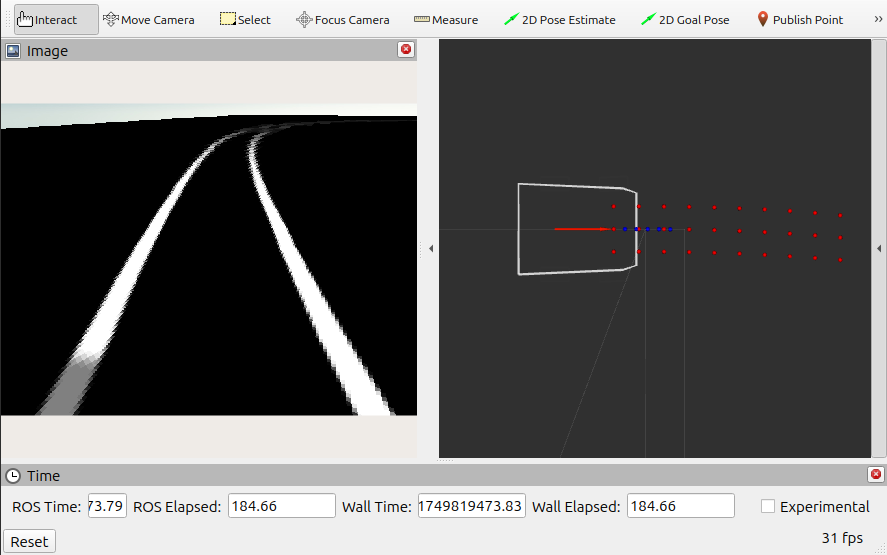}
    \caption{RViz showing debugging information}
    \label{Fig:2.17}
\end{figure}

\begin{figure}[H]
    \centering
    \includegraphics[width=0.75\linewidth]{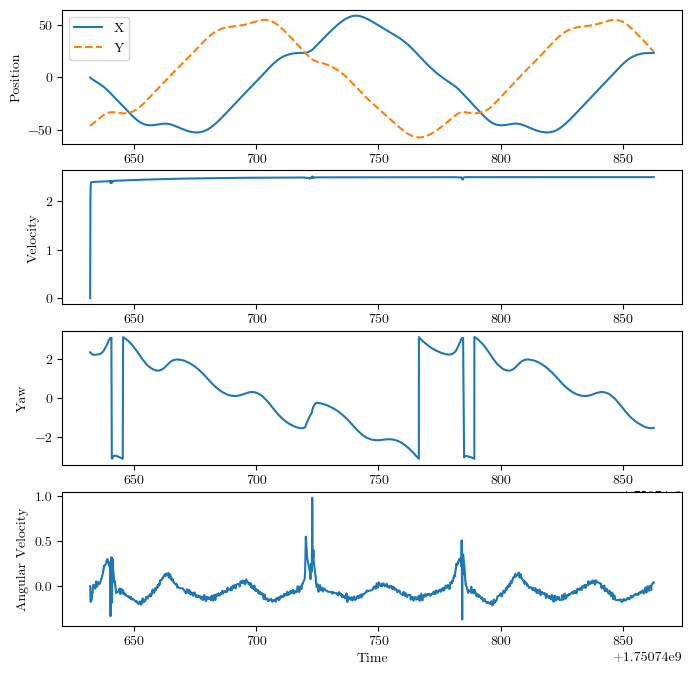}
    \caption{Sample Runtime Information}
    \label{Fig:SampleRun}
\end{figure}

\section{RESULTS \& OBSERVATION}

The experiment was conducted across multiple trials using randomly generated tracks of similar complexity, created by adding sinusoidal curvatures to a uniform circular path, as illustrated in Fig. \ref{Fig:2.19}. The initial velocity profiles of the vehicle for PID and MPC controllers are depicted in Fig. \ref{Fig:2.20}. Overall, both controllers demonstrated comparable performance; however, they each exhibit distinct operational characteristics. The PID controller maintained an almost constant velocity throughout each run, including sharp turns. The behavior produced apparent spikes in angular velocity, as shown in Fig. \ref{Fig:2.21}. In contrast, MPC produced smoother and more regulated control actions. Its adherence to state and input constraints, however, introduced oscillations in angular velocity (Fig. \ref{Fig:2.22}) and tracking error. Consequently, PID achieves lower trajectory deviation, while MPC provides more robust, constraint-aware control at the expense of some accuracy, as summarized in Table \ref{tab:01}.

\begin{table}[htbp]
\caption{Mean Squared Deviation from all of the runs}
\begin{center}
\begin{tabular}{|c|c|c|}
\hline
\textbf{Metric} & \textbf{PID} & \textbf{MPC} \\
\hline
Lateral Mean Squared Deviation & 0.0136 $m^2$ & 0.0390 $m^2$ \\
\hline
Angular Mean Squared Deviation & 0.000548 $rad^2$ & 0.001014 $rad^2$ \\
\hline
\end{tabular}
\label{tab:01}
\end{center}
\end{table}

\begin{figure}[!htb]
    \centering
    \begin{minipage}{0.48\linewidth}
        \centering
        \includegraphics[width=0.95\linewidth]{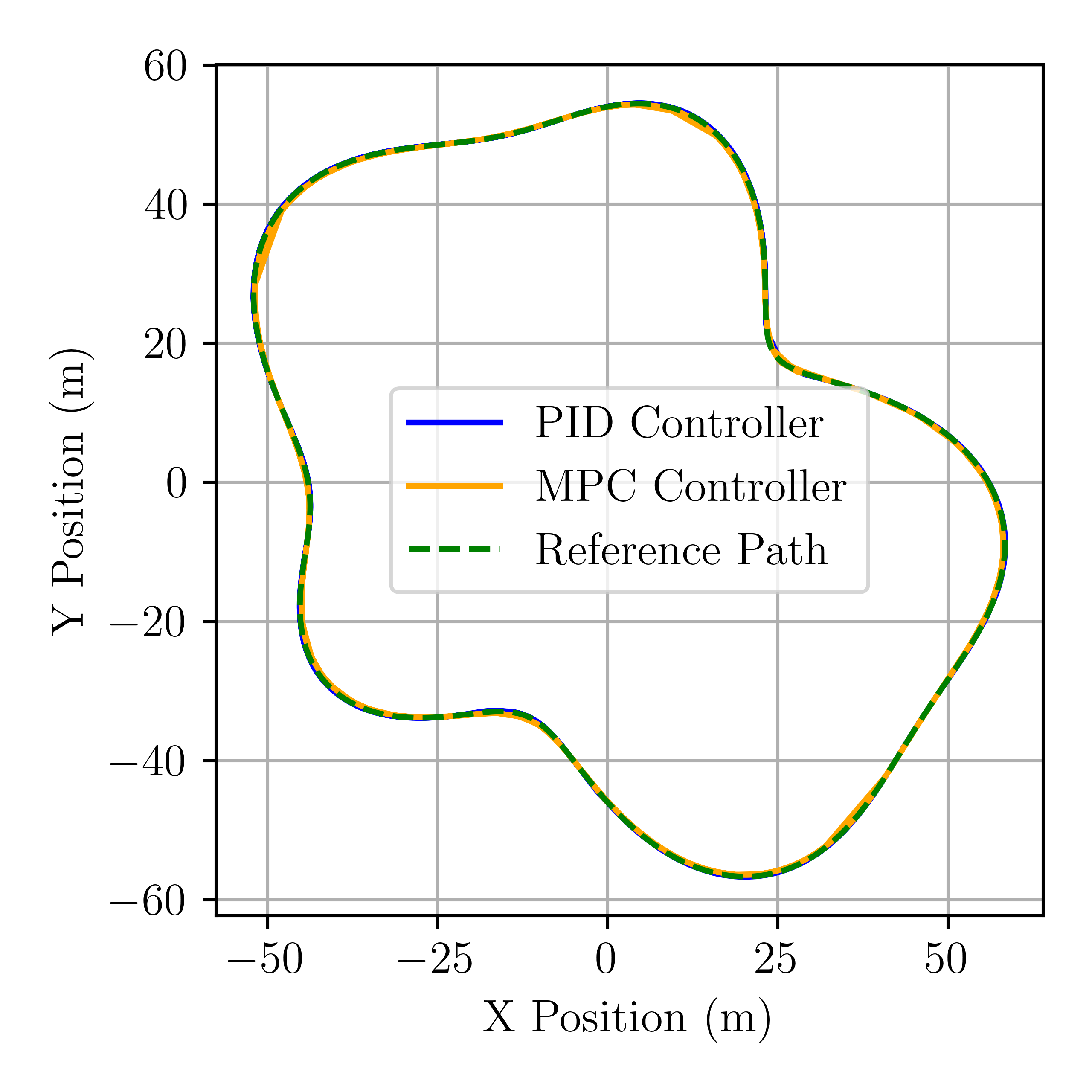}
        \caption{Vehicle Trajectories in every run}
        \label{Fig:2.19}
    \end{minipage}%
    \hfill
    \begin{minipage}{0.48\linewidth}
        \centering
        \includegraphics[width=0.95\linewidth]{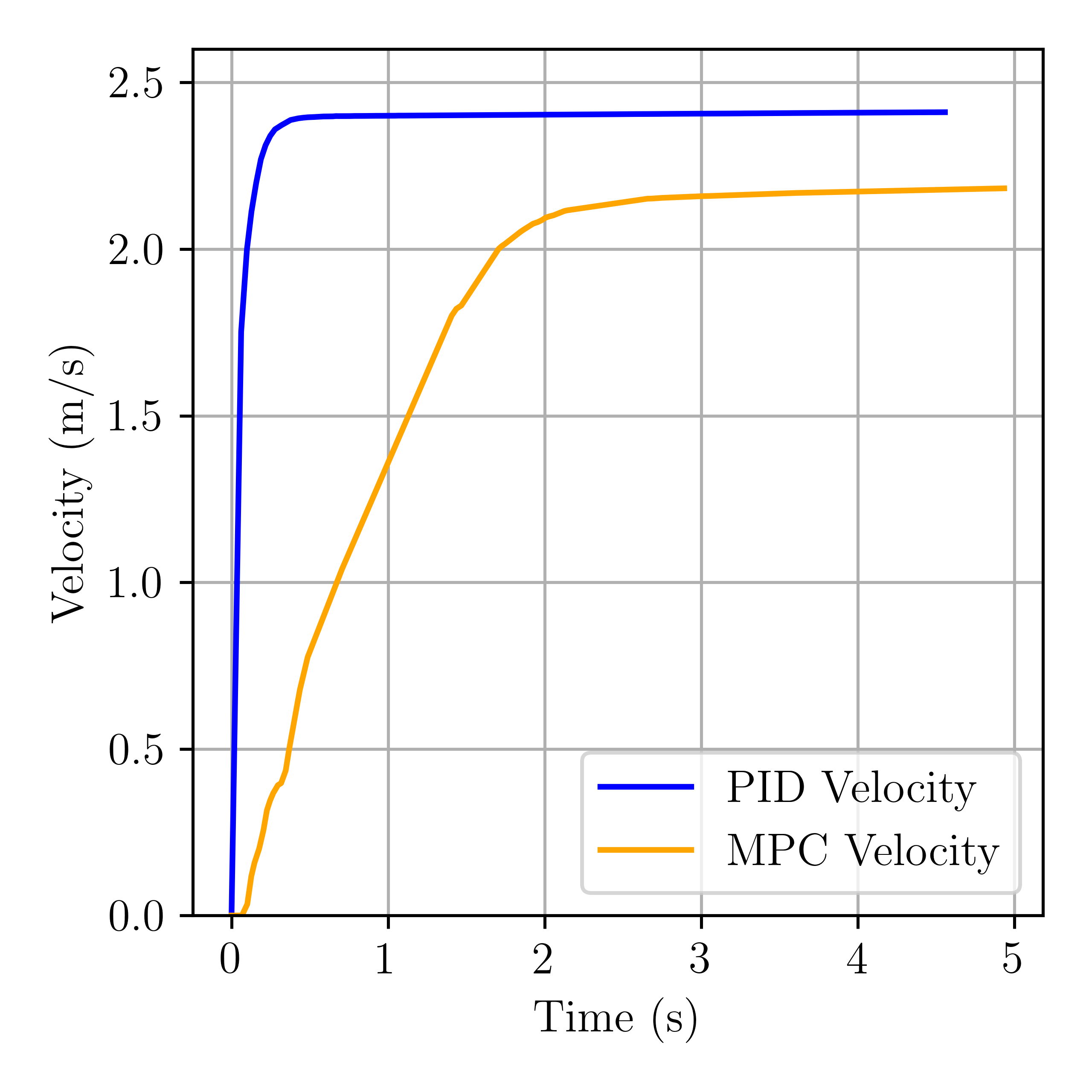}
        \caption{Initial vehicle velocities}
        \label{Fig:2.20}
    \end{minipage}
\end{figure}

\begin{figure}[!htb]
    \centering
    \begin{minipage}{0.48\linewidth}
        \centering
        \includegraphics[width=0.95\linewidth]{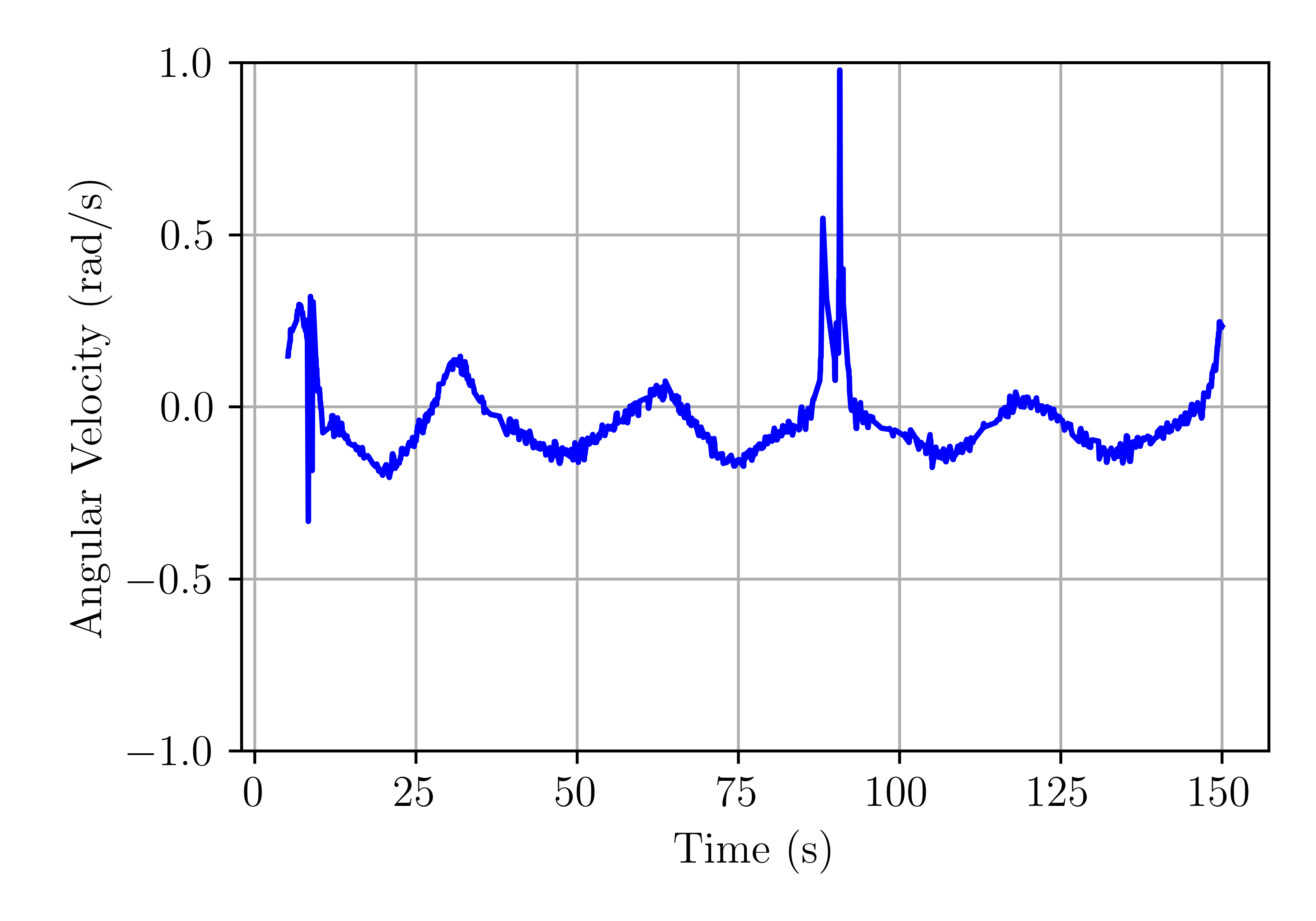}
        \caption{Angular Velocity - PID}
        \label{Fig:2.21}
    \end{minipage}%
    \hfill
    \begin{minipage}{0.48\linewidth}
        \centering
        \includegraphics[width=0.95\linewidth]{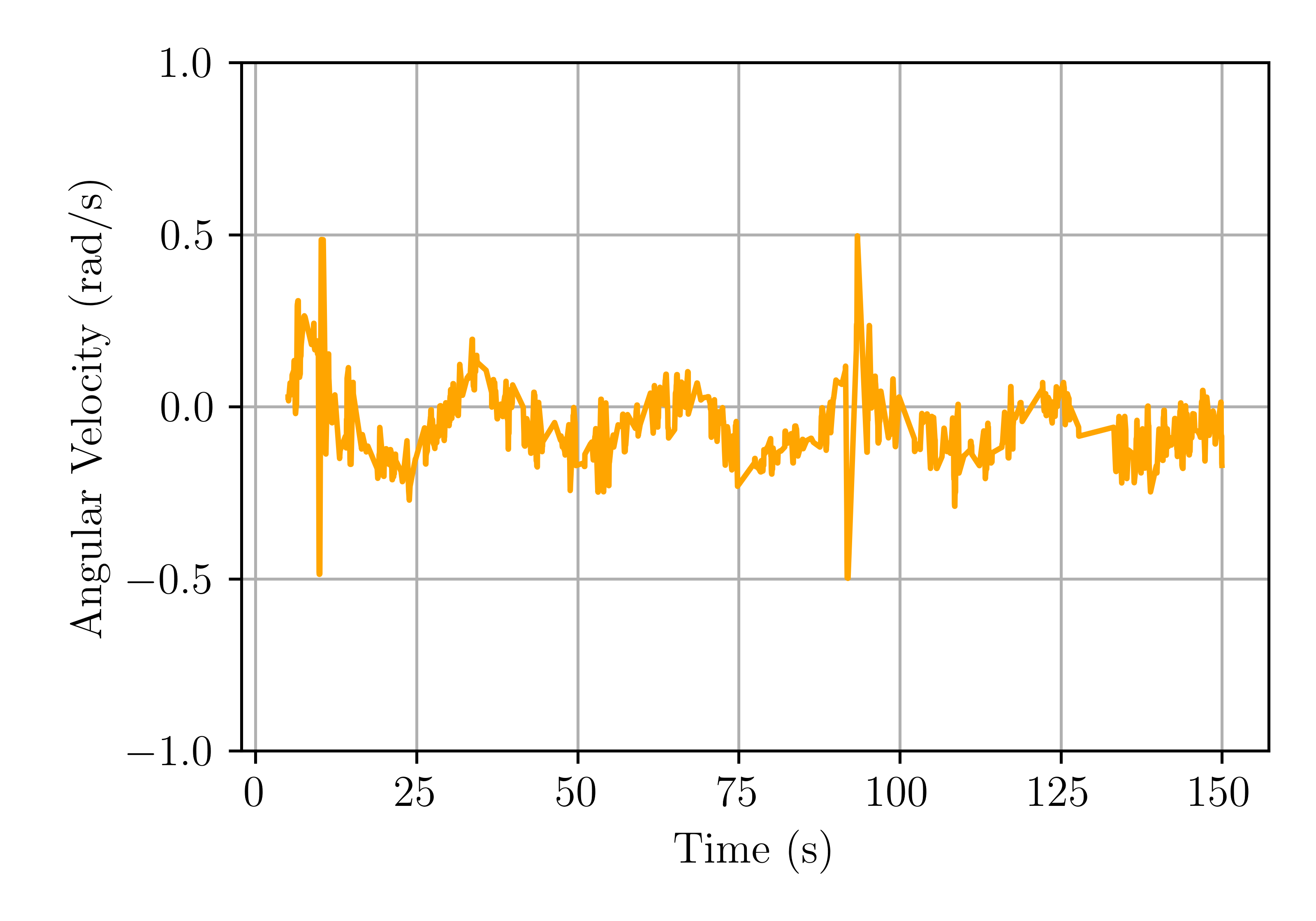}
        \caption{Angular velocity - MPC}
        \label{Fig:2.22}
    \end{minipage}
\end{figure}

\section*{Conclusions}
This work presents a low-cost, open-sourced SIL framework for Ackermann-steered autonomous vehicles by incorporating monocular vision, real-time telemetry, and flexible control experimentation. The platform allows for reliable testing of lane-tracking algorithms while maintaining low computational overhead.

Experiments with PID and MPC controllers show that the testbed successfully reflects the differences in control behavior: PID has a smaller tracking error, whereas MPC offers much smoother, constraint-aware control. The results presented herein validate this framework as a practical means of developing and comparing algorithms.

The overall system is set up to be easily extensible for future studies in multi-agent scenarios, learning perception, and onward development into Hardware-in-the-Loop tests.

\bibliographystyle{unsrt}
\bibliography{bibfile}
\end{document}